\documentclass{article}

\usepackage[preprint]{neurips_2025}

\usepackage[utf8]{inputenc} % allow utf-8 input
\usepackage[T1]{fontenc}    % use 8-bit T1 fonts
\usepackage{hyperref}       % hyperlinks
\usepackage{url}            % simple URL typesetting
\usepackage{booktabs}       % professional-quality tables
\usepackage{amsfonts}       % blackboard math symbols
\usepackage{nicefrac}       % compact symbols for 1/2, etc.
\usepackage{microtype}      % microtypography
\usepackage{xcolor}         % colors
\usepackage{amsthm}
\usepackage{lipsum}
\usepackage{enumitem}
\usepackage{graphicx}
\newtheorem{theorem}{Theorem}[section]

\theoremstyle{definition}
\newtheorem{definition}[theorem]{Definition}

\hypersetup{
 colorlinks=true,
 citecolor=blue,
 linkcolor=red,
 urlcolor=blue
 }
 
\title{A Call for Collaborative Intelligence: Why Human-Agent Systems Should Precede AI Autonomy} % consider something like

\author{Henry Peng Zou\textsuperscript{1*},
Wei-Chieh Huang\textsuperscript{1*},
Yaozu Wu\textsuperscript{2*},
Chunyu Miao\textsuperscript{1},  \\ \bf
Dongyuan Li\textsuperscript{2\dag},
Aiwei Liu\textsuperscript{3}, 
Yue Zhou\textsuperscript{1}, 
Yankai Chen\textsuperscript{1\dag}, 
\textbf{Weizhi Zhang}\textsuperscript{1},  \\ \bf
\textbf{Yangning Li}\textsuperscript{3},
\textbf{Liancheng Fang}\textsuperscript{1},
\textbf{Renhe Jiang}\textsuperscript{2},
\textbf{Philip S. Yu}\textsuperscript{1}
\\
\textsuperscript{1}University of Illinois Chicago,
\textsuperscript{2}University of Tokyo,
\textsuperscript{3}Tsinghua University\\
\texttt{\{pzou3, whuang80, psyu\}@uic.edu}, \texttt{yaozuwu279@gmail.com}, \\
\texttt{lidy@csis.u-tokyo.ac.jp},
\texttt{yankaichen@acm.org}
}

\begin{document}

\maketitle
\renewcommand{\thefootnote}{}\footnote{$^*$ Equal Contribution. $^\dag$ Corresponding Authors.}

\begin{abstract}
Recent improvements in large language models (LLMs) have led many researchers to focus on building fully autonomous AI agents. This position paper questions whether this approach is the right path forward, as these autonomous systems still have problems with reliability, transparency, and understanding the actual requirements of human.
We suggest a different approach: LLM-based Human-Agent Systems (LLM-HAS), where AI works with humans rather than replacing them. By keeping human involved to provide guidance, answer questions, and maintain control, these systems can be more trustworthy and adaptable.
Looking at examples from healthcare, finance, and software development, we show how human-AI teamwork can handle complex tasks better than AI working alone. We also discuss the challenges of building these collaborative systems and offer practical solutions.
This paper argues that progress in AI should not be measured by how independent systems become, but by how well they can work with humans. The most promising future for AI is not in systems that take over human roles, but in those that enhance human capabilities through meaningful partnership.

\end{abstract}

\section{Introduction}

AI assistants capable of independent operation have long captivated imagination and scientific pursuit, spanning from speculative fiction to early research on autonomous problem-solving systems~\citep{dautenhahn1998art,gizzi2022creative}. Recent advancements in Large Language Models (LLMs) have rekindled this foundational vision with unprecedented success: LLM-based autonomous agents promise to achieve complex goals with the ability to perceive, plan, and act in dynamic environments, with minimal human intervention~\citep{wang2024survey}.
This rapid technological progress has naturally led many researchers and practitioners toward an ``autonomy-first'' mindset~\citep{cihon2025measuring}. The prevailing assumption is that more autonomous agents are better—that reducing human involvement is inherently desirable and that complete independence should be the ultimate goal. Industry leaders and academic institutions have invested heavily in pursuing systems that can operate with minimal human oversight, driven by visions of AI agents that can handle entire workflows from start to finish~\citep{ferrag2025llm, hu2025unified}.
However, this rush toward full autonomy, while understandable given recent breakthroughs, may be \textit{premature} \cite{deng2024deconstructing,mitchell2025fully,zou2025survey}.

\textbf{Our position} is that \textbf{deploying fully autonomous LLM-based agents in complex real-world scenarios at this stage of development poses significant risks and limitations that could undermine both safety and effectiveness}. Rather than viewing autonomy as the primary measure of progress, we argue for a fundamental shift toward LLM-based Human-Agent Systems (LLM-HAS) \cite{zou2025survey}, where AI agents function as active teammates rather than independent operators. 
We advocate for the focused development and deployment of collaborative, supportive, ethical, and adaptive AI-human partnerships that enhance human capabilities while maintaining essential human oversight and judgment. 
This approach does not represent a retreat from AI's ambitious goals, but rather a redefinition of what constitutes advanced AI—measuring progress not by isolation, but by collaborative intelligence.

\begin{figure*}[!t]
  \centering
     \includegraphics[width=0.95\textwidth]{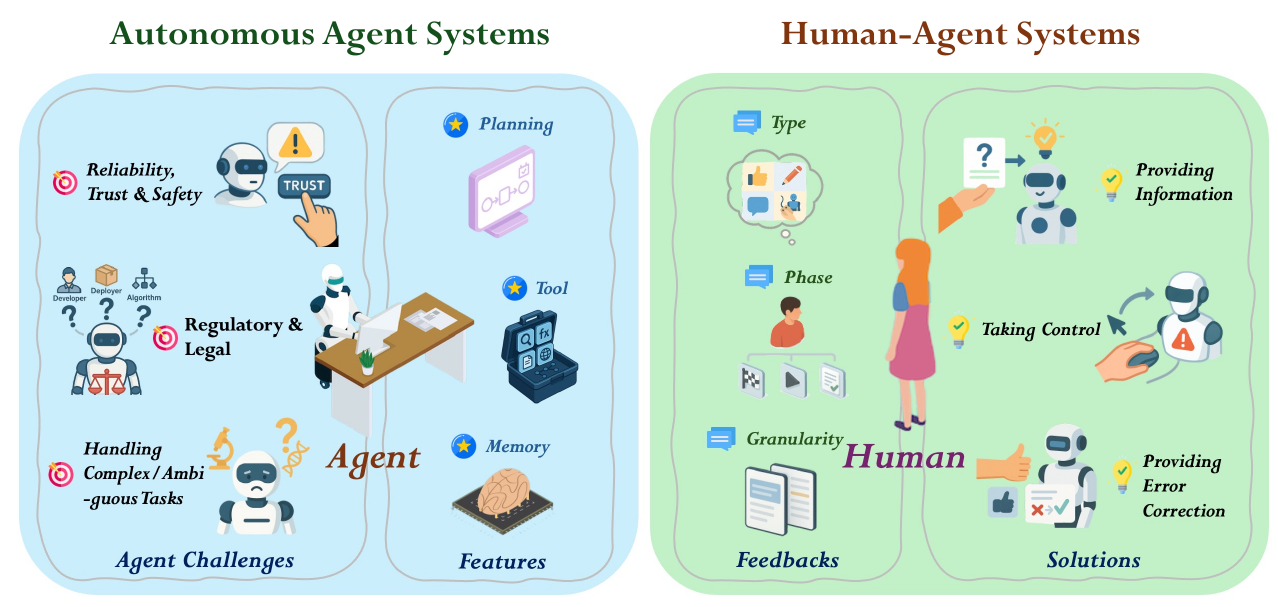}
    % {./figures/motivation_llm_has.pdf}
  \caption{From Autonomous Agent Systems to Human-Agent Systems.}
  \label{fig:overview_llm_has}
\end{figure*}

  % \caption{From Fully Autonomous Agent to Human-Agent System.}

This paper advocates a fundamental shift from pursuing autonomous LLM agents to prioritizing collaborative HAS. We begin by critically examining the current trajectory towards fully autonomous LLM-based agents, identifying limitations in reliability, complexity handling, and ethical concerns (Section~\ref{Section-2}). Building on these identified shortcomings, we present LLM-based HAS as a paradigmatic alternative, establishing foundational principles that demonstrate how human-LLM collaboration directly addresses the core weaknesses of fully autonomous approaches (Section~\ref{Section-3}). To substantiate our position, we highlight the emerging significance of HAS across multiple domains, showcasing promising results (Section~\ref{Section-4}). We acknowledge that HAS approaches face their own challenges; Therefore, we list these key limitations and propose concrete research directions to address these issues (Sections~\ref{Section-5} \& \ref{Section-6}). Finally, we present Alternative Views in Section~\ref{Section-7}.

\section{Existing Pursuit: Autonomous LLM-based Agents}\label{Section-2}

\begin{definition}[LLM-based Autonomous Agent]
An \textit{LLM-based autonomous agent} is a system that operates independently in open-ended real-world environments by completing tasks through a perception-reasoning-action loop without human intervention ~\cite{wang2024survey, xie2024large}.
\end{definition}

 Unlike human-in-the-loop systems, LLM-based Autonomous Agent interprets goals, plans actions, invokes tools, and adapts using language-based reasoning and memory — all autonomously.

To concretize the definition, we highlight real-world deployments that showcase the autonomous capabilities of such agents. \textbf{(1) In software engineering}, GitHub Copilot\footnote{\url{https://github.com/features/copilot}} exemplifies how agents can autonomously generate, test, and refactor code with minimal developer input, significantly accelerating routine development workflows~\cite{liu2024large}. \textbf{(2) In customer support}, systems like Manus\footnote{\url{https://manus.im/}} and Genspark\footnote{\url{https://www.genspark.ai/}} perform complex itinerary planning, automate bookings, and resolve service issues without human oversight, demonstrating robust perception-action loops in dynamic environments.  Overall, these applications highlight a compelling vision: frictionless deployment, continuous 24/7 operation, and easy scalability will fundamentally reshape the possibilities of automation in industries, research, and daily life~\cite{shen2025mind}.

However, current LLM-based autonomous agents face significant challenges in real-world deployment:
\textbf{1. Reliability, Trust, and Safety.} LLMs generate “hallucination” outputs that appear plausible but are in fact completely fabricated~\cite{huang2025survey}.
The prevalence of hallucinations directly undermines trust in fully autonomous agents. If an autonomous system cannot consistently and reliably provide accurate information, its utility in high-risk environments (such as medical diagnosis~\cite{qiu2024llm}, financial decision-making~\cite{yu2024fincon}, or critical infrastructure control) will be severely compromised. 
\textbf{2. Handling Complex and Ambiguous Tasks.} Agents struggle with tasks requiring deep reasoning, particularly when goals are ambiguous. Human instructions are often underspecified; without commonsense context, agents can misinterpret them and take incorrect actions. This makes them unreliable for complex domains like scientific research, where objectives are frequently open-ended.
\textbf{3. Regulatory and Legal Challenges.} Fully autonomous agents, despite their ability to act, are not formally accountable under existing law~\cite{ong2024ethical}. This ambiguity creates a large accountability and transparency gap. When they cause harm or make an incorrect decision, it becomes extremely difficult to determine responsibility~\cite{holbrook2024overtrust}. Is it the developer, the deployer, or the algorithm itself? As the capabilities of intelligent agents increase, the legal gap between ``capabilities'' and ``obligations'' grows wider.

\section{Towards LLM-Based Human-Agent Systems}\label{Section-3}

The persistent challenges in the pursuit of fully autonomous LLM-based agents—spanning safety, ethics, reliability, and complexity—necessitate a practical shift in paradigms. Instead of isolating human involvement, the LLM-HAS paradigm takes advantage of human strengths for creating more robust, effective, and trustworthy systems (as shown in Figure~\ref{fig:overview_llm_has}).

\subsection{LLM-based Human-Agent Systems}

\begin{definition}[LLM-based Human-Agent Systems]
An \textit{LLM-based Human-Agent System} is a collaborative framework where humans and LLM-powered agents interact to accomplish tasks. 
\end{definition} 

Unlike fully autonomous agents, these systems maintain humans in the loop to provide critical information and clarifications \citep{kim2025beyond,naik2025empirical,zou2025survey}, offer feedback by evaluating outputs and guiding adjustments \citep{gao2024taxonomy, dutta2024problem, li2024fb}, and assume control in high-stakes or sensitive scenarios \citep{chen2025reinforcing, natarajan2025human, xiao2023llm}. 
This human involvement in LLM-HAS ensures enhanced performance, reliability, safety, and explicit accountability, particularly where human judgment remains indispensable.

% \subsection{Advantages/Significance and Potential of LLM-HAS}
\subsection{Advantages of LLM-HAS}

The advantages and rationale for prioritizing HAS stems directly from its potential to address the critical limitations and risks associated with autonomous agent systems:

\textbf{Improved Trust and Reliability:} The interactive nature of HAS allows humans to provide crucial feedback, correct potential LLM hallucinations in real-time~\citep{yang2023humanintheloopmachinetranslationlarge}, verify information, and guide the agent toward more accurate and reliable outputs~\citep{10.1145/3491102.3517582}. This collaborative verification process is essential for building trust, especially where the cost of error is high.

\textbf{Managing Complexity and Ambiguity.} Unlike autonomous agents that struggle with unclear instructions, LLM-HAS excels through continuous human clarification~\cite{geissler2024concept}. Humans provide essential context, domain expertise, and progressive refinement of ambiguous goals—critical capabilities for complex tasks. When faced with an underspecified objective, the system can request clarification rather than proceeding with potentially incorrect assumptions~\cite{miao2025clarigen}, making LLM-HAS particularly effective for open-ended research or creative endeavors where objectives evolve dynamically \citep{du-etal-2024-llms}.

\textbf{Clearer Lines of Accountability:} With a human involved in the decision-making process, especially in supervisory or interventional roles, establishing accountability becomes more straightforward. The human operator or supervisor can often be designated the responsible party, simplifying the legal and regulatory landscape compared to situations where an autonomous agent makes a critical error~\cite{yang2024supercorrect}.

% \newpage
\section{Applications of LLM-powered Human-Agent Systems}\label{Section-4}

LLM-HAS are increasingly used in domains that rely heavily on human input, contextual reasoning, and real-time interaction~\citep{han2025convcodeworld, zhou2025sweetrltrainingmultiturnllm}.
Their core advantage lies in treating LLMs not as passive language generators but as active partners that can understand user goals, plan actions, and adjust their behavior through ongoing dialogue.
This iterative communication helps align agent behavior with human intent, making collaboration more flexible, transparent, and effective than traditional rule-based or end-to-end systems~\citep{wang2025survey,seo2025reveca,zhang2025leveraging}.

Specifically, in \textbf{Embodied AI}, LLM-HAS enables agents to follow complex instructions and coordinate physical tasks with humans using natural language~\citep{chang2024partnr,tanneberg2024help}. 
In \textbf{Software Development}, agentic assistants support multi-turn problem-solving, integrating human feedback to refine and adapt code generation~\citep{feng-etal-2024-large,wang2024mint}. 
\textbf{Conversational Systems} benefit from agents that proactively ask clarifying questions, plan dialogue, and explain reasoning steps, improving controllability and user trust~\citep{shao2024collaborative,pan2024agentcoord}. 
In \textbf{Gaming}, LLM-HAS facilitates dynamic cooperation, allowing real-time collaboration with human players with uncertain or evolving objectives~\citep{liu2023llm,mehta-etal-2024-improving}. 
In \textbf{Finance}, collaborative HAS like FinArena demonstrate how pairing LLM agents with experienced investors can enhance market prediction and portfolio performance~\citep{xu2025finarena}.
In \textbf{Healthcare}, LLM-HAS support both patient-facing services and clinical workflows, facilitating diagnosis, treatment planning, and drug discovery through seamless integration of medical expertise and language-based reasoning~\citep{li2024mmedagent,patel2025ai}. 
In \textbf{Autonomous Driving}, they enable intent-aware driving assistance, adaptive feedback loops, and shared control models~\citep{cui2024drive,wu2025multi}. 
Across these domains, a common methodological pattern emerges: LLM-HAS redefines human-AI interaction as a collaborative process based on language, shaped by feedback, and driven by adaptive reasoning. 
This unified paradigm raises key questions about alignment, transparency, and co-adaptation, opening valuable space for discussion within the AI community.  
As these systems advance, they bring both significant promise and serious challenges, requiring careful study of their technical design and broader social impact~\citep{chen2024end,ma2024learning}.

% \newpage
\section{Key Challenges in Human-Agent Systems}\label{Section-5}

While LLM-HAS represents a promising direction, its successful implementation requires careful consideration of several inherent challenges throughout its development lifecycle. We discuss these limitations and propose potential solutions and future research directions \citep{zou2025survey}.

\noindent \textbf{[Initial Setup] Mostly Agent-Centered Work.}  
Most current research on LLM-HAS adopts an agent-centered view, where humans primarily evaluate agent outputs and provide corrective feedback~\citep{zou2025survey}. 
This unidirectional interaction dominates existing paradigms. 
However, there is a compelling opportunity to reshape this dynamic. 
Enabling agents to actively monitor human performance, detect inefficiencies, and offer timely suggestions would allow their intelligence to be used effectively and reduce human workload~\citep{liang2024learningcooperatehumansusing}. 
When agents transition into an instructive role—proposing alternative strategies, highlighting potential risks, and reinforcing best practices in real-time—both human and agent performance improves. We argue that shifting toward a more human-centered or equitable LLM-HAS design is essential to fully realize true human-agent teamwork~\citep{klieger2024chatcollabexploringcollaborationhumans}.

\noindent \textbf{[Data] Human Flexibility and Variability. } 
Human feedback in LLM-HAS varies significantly in role, timing, and style~\cite{fernandes-etal-2023-bridging}. 
Since humans are subjective and influenced by their personalities, different individuals can lead to diverse outcomes when interacting with the same LLM-HAS. This highlights a crucial need: first, for thorough investigations or benchmarks into how varied human feedback affects entire systems; and second, for flexible frameworks that can adopt to this diversity~\cite{zou2025survey}. 
Moreover, humans are often under-evaluated in LLM-HAS, creating an imbalance that may obscure whether performance bottlenecks stem from the agent or the human~\citep{feng2024largelanguagemodelbasedhumanagent}. Resolving this requires fair interaction protocols and shared evaluation standards.
Another challenge lies in the widespread use of LLM-simulated human proxies, which often fail to reflect the variability of real human input, introducing performance gaps that undermine the validity of comparisons~\citep{wang2024mint, han2025convcodeworldbenchmarkingconversationalcode}.

% Optional
\noindent \textbf{[Model Engineering] Lacking Adaptivity and Continuous Improvement.} 
A core challenge in LLM-HAS development is building truly adaptive and continuously improving AI teammates. Previous approaches treat LLMs as fixed, pre-trained tools, thereby missing opportunities for dynamic evolution within collaborative settings~\cite{luo2025large}. This static view introduces three key challenges. First, most systems fail to adequately leverage human insights. Without advanced ways to incorporate diverse human guidance (e.g., preferences, critiques), LLMs struggle to become genuinely teachable and context-aware~\cite{yang2025lighthouse, asai2023self}. 
Second, models lack robust capacity for continual learning and knowledge retention in dynamic environments. This prevents them from building long-term expertise and can lead to catastrophic forgetting, severely hindering their growth as collaborators~\cite{jin2025exploring,li2025llm}. 
Third, the absence of real-time optimization—such as adaptive prompting and self-correction—hampers efficiency, alignment, and resource use~\cite{subramonyam2025prototyping,upadhyaya2024internalized}. 
Effectively addressing these challenges through integrated human feedback, lifelong learning, and dynamic optimization is key to unlocking the full potential of LLM-HAS in human-agent collaboration.

\noindent \textbf{[Development] Unresolved Safety Vulnerabilities.} 
LLM-HAS face critical challenges in sustaining safety, robustness, and accountability post-deployment. While performance often takes precedence, crucial aspects like reliability, security, and user privacy in human interaction remain underexplored \citep{qiu2025emoagent}. This leads to three key issues: 
First, without robust monitoring and continuous evaluation, systems risk undetected misaligned behaviors, unpredictable failures, or unintended data disclosures in real-time \citep{zhang2025characterizing}. 
This lack of vigilance hinders the rapid iteration guided by humans.
Second, inadequate alignment maintenance and human oversight make dynamic LLM-HAS vulnerable to unpredictable agent adaptation and behavioral drift, complicating liability for delegated actions without clear attribution~\citep{chan2025infrastructure}.
Finally, overlooking long-term adaptation and responsible AI undermines safety and trust. Ensuring reliable human-agent collaboration requires ongoing monitoring, strong oversight, and integrated responsible AI practices~\citep{chojnacki2025interpretable}.

\noindent \textbf{[Evaluation] Inadequate Evaluation Methodologies.} 
Existing evaluation frameworks for LLM-HAS are fundamentally flawed. They primarily emphasize agent accuracy and static benchmarks, often entirely ignoring the real burden placed on human collaborators \citep{ma2025sphere}. 
This oversight means crucial aspects remain unmeasured: 
First, standard metrics for human workload and efficiency are lacking. Humans invest varying time and effort to feedback, yet this critical ``cost'' remains unsystematically quantified. Evaluations must extend beyond mere output accuracy to cover factors like feedback time across all collaboration phases \citep{chen2025advancinghumanmachineteamingconcepts}. 
Second, as human expertise and LLM-based agent capabilities merge, uncertainty and variability grow. Current evaluations fail to capture nuances of interaction quality, the dynamics of trust, transparency, and explainability, or adherence to ethical alignment and safety beyond simple performance. Moreover, overall user experience and cognitive load are rarely assessed holistically. 
A new evaluation approach or set of metrics comprehensively quantifying contributions and costs for both humans and agents across these critical dimensions is essential for truly efficient and responsible collaboration.

% \newpage
\section{Human-Agent Systems Implementation Guidelines}\label{Section-6}
%% Lifecycle
%% Implementation Guideline <- from the aspect 'Potential Solution'

\begin{figure*}[!t]
  \centering
\includegraphics[width=\textwidth]{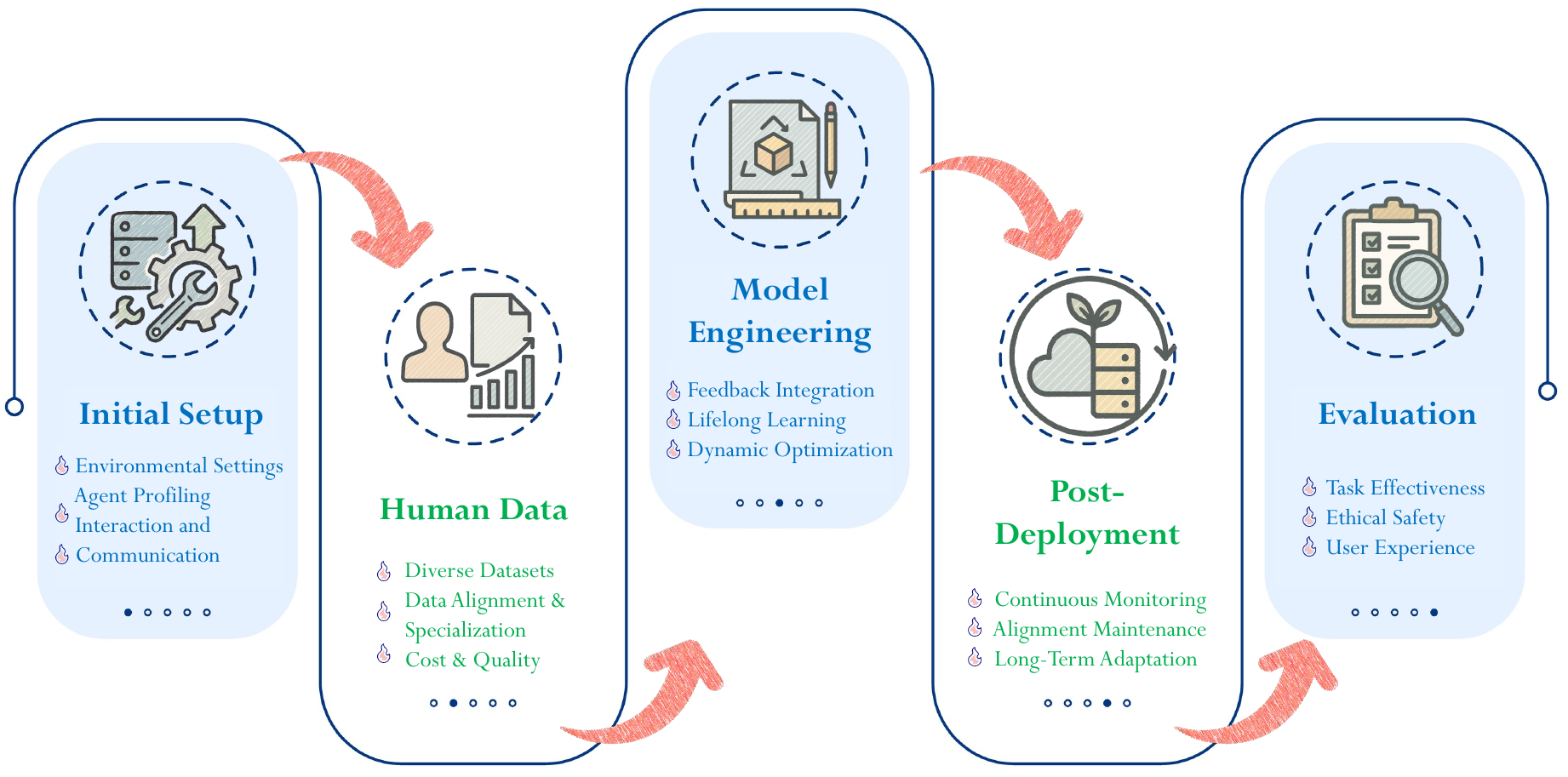}
  \caption{Implementation guidelines for the Human-Agent Systems. More details in Section \ref{Section-6}.
  }
  \label{fig:implementation}
\end{figure*}

Implementing an effective LLM-HAS requires a systematic framework in which every component is clearly defined and seamlessly integrated. This framework are partitioned five key domains, as shown in Figure~\ref{fig:implementation}: (1) \textbf{Initial setup}, where \textit{interaction paradigms}, \textit{human feedback phases} and \textit{interaction architecture} for human–agent collaboration are specified; (2) \textbf{Human Data}, the proper data for LLM-HAS; (3) \textbf{Model Engineering}, an iterative process of \textit{fine-tuning}, \textit{modular design}, and \textit{continuous optimization} that enhances flexibility, adaptability, and alignment with user needs; (4) \textbf{Post-Deployment Evaluation and Monitoring}, involving \textit{continuous performance assessment}, \textit{human-in-the-loop feedback loops}, and governance mechanisms to guarantee reliability, ethical compliance, and the long-term co-evolution of the human-agent partnership; and (5) \textbf{Evaluation} in different stages of the LLM-HAS.

\subsection{Initial Setup: Architecture for Collaboration}

The initial setup phase for implementing LLM-HAS is crucial for the whole task and system. It requires careful definition of the environment, clear profiling of human and agent roles and capabilities, the design and architecture of interaction, and strategic configuration of the LLM's core functionalities, including knowledge grounding and tool integration.

\textbf{Environmental Settings.} The environmental settings and profiling define where the interaction between the human and the agent occurs and the internal status of both the human and the agent. It involves defining the \textbf{shared interaction space} \textit{(physical or virtual)} \citep{zou2025survey} and \textbf{profiling human users} \textit{("lazy" vs. "informative")} \citep{wang2023mint} alongside \textbf{agent roles} \textit{(assistant, specialist, or RACI (Responsible, Accountable, Consulted, Informed))} \citep{talebirad2023multi, smith2005role,samuel2024personagym} and capabilities like planning and memory \citep{zhang2024survey}. LLM-based agents possess a degree of interpretive flexibility not found in deterministic model. Therefore, establishing clear environmental configurations and thorough profiling is essential to prevent inefficiencies in LLM-HAS that typically arise from ambiguous roles and system design.

\textbf{Interaction and Communication.} Interaction types should be specified at a fine-grained level. Rather than broadly characterizing collaboration, human must explicitly distinguish between \textit{supervision}, \textit{delegation}, \textit{cooperation}, and \textit{coordination}. Additionally, the \textbf{orchestration strategy} (\textit{one-by-one, simultaneous}) and \textbf{synchronization mode} (\textit{synchronous, asynchronous}) must be defined to clarify how humans and LLM-based agents interact \citep{zou2025survey}. Also, the human feedback phase (\textit{initial-setup}, \textit{during task}, \textit{post task}) and granularity (\textit{holistic}, \textit{segment-level}) must also be well-defined. In terms of Interaction protocols, they should be grounded in communication theories (e.g., Gricean Maxims) \citep{kim2025applying} to foster cooperative interactions and situational awareness \cite{ cheng2025exploratory}, enabling AI systems to build a shared understanding with human collaborators.  

\subsection{Characterizing Human Data}

The foundation of a well-aligned and effective LLM-HAS lies in the strategic acquisition, processing, and use of diverse high-quality data, especially human-generated data. This is crucial for equipping agents with nuanced understanding, enhancing their collaborative capabilities, and ensuring they align with human preferences and values.

\textbf{Diverse Datasets.} Effective HAS implementation uses various human data, including interaction logs, to understand real-world user behavior \citep{kyrychenko2025human}. It also incorporates both explicit feedback (e.g., \textit{corrections}, \textit{ratings}) and implicit feedback (e.g., \textit{inferred preferences}) from humans to guide agent adaptation \citep{zou2025survey}. In terms of domain, task-specific corpora, such as those related to cybersecurity \citep{tihanyi2024cybermetric}, help LLMs learn how to ensure security. Similarly, datasets like \textit{XtraQA} \citep{chen2025xtragpt} provide decent writing examples for enhancing collaborative academic writing. For multimodal agents, datasets such as \textit{LLaVA-RLHF} \citep{sun2024aligning} and \textit{VLFeedback} \citep{li2024vlfeedback}, which align visual and textual reasoning, provide insight into enhancing collaboration with humans. Considering the variety of the dataset, human must specify it clearly.

\textbf{Alignment \& Specialization.} The increasing specialization of datasets, as demonstrated by BeaverTails \citep{ji2023beavertails}, which specifically annotates \textit{helpfulness} and \textit{harmlessness}, reflects a growing recognition that "alignment" is a multifaceted concept. This complexity requires targeted data strategies that go beyond general preferences. As LLM-HAS becomes more sophisticated, the traditional "one-size-fits-all" data approach is proving to be less effective. This trend suggests that more detailed datasets will emerge, and frameworks must be developed to manage them effectively.

\textbf{Cost \& Quality.} The high cost of human annotation \citep{yu2025aligning} is prompting innovation in hybrid data generation. This approach combines data collected by humans with synthetic data generated by LLMs \citep{borchers2025augmenting}. While this method can improve the scalability and extendability of data collection, it also risks perpetuating biases present in flawed "teacher" LLMs. Therefore, it is crucial to implement robust validation processes and maintain human oversight to prevent \textit{"model inbreeding"} and ensure that the quality of the data aligns with real-world human experiences.

\subsection{Model Engineering: Iterative Development for Adaptive Teammates}

The model engineering in LLM-HAS development is inherently iterative and adaptive, encompassing three core dimensions: \textbf{(1) seamless integration of human feedback}, \textbf{(2) Lifelong Learning}, and \textbf{(3) dynamic optimization} via continuous refinement and self-correction.

\textbf{Integration of Human Feedback.} The iterative integration of human feedback through techniques such as reinforcement learning from human feedback (RLHF) \citep{wang2023rlhf, du2025survey}, reinforcement learning from AI feedback (RLAIF) \citep{lee2023rlaif}, direct preference optimization (DPO) \citep{ivison2024unpacking}, and critique-guided improvement (CGI) \citep{yang2025lighthouse, asai2023self} can help LLM-based agents improve their performance. The LLM-HAS implementations shall apply methods to transform static, pre-trained LLMs into adaptive, teachable agents by leveraging human-ranked responses, LLM-based critiques, and natural-language guidance to drive continuous refinement.

\textbf{Lifelong Learning.} LLM-based agents must be designed for lifelong learning, continuously adapting in dynamic environments by acquiring and retaining knowledge without catastrophic forgetting \citep{zheng2025lifelong}, supported by memory mechanisms like InfLLM for processing long sequences \citep{xiaoinfllm}. This pursuit aims to build agents that accumulate experience and develop a form of "expertise" over time. Furthermore, teachable agents that learn through human instruction, constructed based on paradigms such as Learning by Teaching \citep{jin2025exploring} and interactive human-in-the-loop frameworks like GradeHITL \citep{li2025llm}, can use human insights to offer more interpretable feedback. %can capitalize on human strengths in explanation and critique to deliver more interpretable feedback.

\textbf{Dynamic Optimization.} Model engineering involves employing dynamic refinement techniques to enhance agent performance, robustness, and alignment with human collaborators. Dynamic prompt engineering, involving iterative refinement by multi-stakeholder teams \citep{subramonyam2025prototyping}, and self-correction mechanisms, where agents identify and rectify their errors (e.g., ToolMaker \citep{wolflein2025llm}, InSeC \citep{upadhyaya2024internalized}). The system must adopt an optimization strategy to significantly reduce the redundant cost of resources and time to benefit the agent's iteration. 

\subsection{Post-Development: Sustaining and Evolving Human-Agent Partnerships}
After the deployment, LLM-HAS demands continuous vigilance through robust monitoring systems, proactive strategies to maintain alignment and mitigate behavioral drift, clearly defined human oversight mechanisms, and long-term evaluation and updating protocols to ensure sustained performance, safety, and ethical operation.

\textbf{Continuous Monitoring \& Evaluation.} Post-deployment requires continuous vigilance through robust MLOps and AIOps systems that evaluate outputs in real time for reliability, hallucinations, user satisfaction, and task failures \citep{zhang2025characterizing, xu2025nuclear}, coupled with adaptive long-term evaluation \citep{dietz2025llm, xia2024evaluation} that merges development, testing, and deployment to detect catastrophic or deceptive behaviors and enable rapid human-guided iteration.

\textbf{Alignment Maintenance \& Oversight.} LLM-HAS are dynamic systems that require continuous governance \citep{tennant2024moral}, as agents can adapt unpredictably. Human oversight is crucial, especially in sensitive situations. Addressing liability from delegated use requires a principal-agent perspective \citep{gabison2025inherent}, together with an agent infrastructure for attribution of actions and remediation of harm \citep{chan2025infrastructure}. The call for ``Law-Following AIs'' \citep{o2025law} highlights societal expectations for compliance, necessitating ``compliance-by-design'' features.

\textbf{Long-Term Adaptation \& Responsible AI.} Ensuring sustained performance, safety, and ethical operation requires effective long-term adaptation and responsible AI practices. Maintenance strategies include optimizing task allocation, managing context and memory \citep{han2024llm}, and potentially using techniques like representation steering for post-deployment modifications \citep{chojnacki2025interpretable}, all within a framework of responsible AI practices \citep{wang2025surveyres, zhou2024large}.

%\vspace{-2mm}

\subsection{Evaluation}

Since LLM-HAS inherently requires human input and real-time adaptation, traditional static benchmark evaluation may be inadequate. We advocate that LLM-HAS should be evaluated in five domains: 

\noindent \textbf{Task Effectiveness and Efficiency.} 
The primary criterion for evaluating LLM-HAS is its effectiveness and efficiency in downstream applications. This includes factors such as system speed, output accuracy and quality, and resource utilization. Metrics that holistically capture these aspects—such as pass@k and major@k—are increasingly being adopted~\citep{chen2021evaluating}.

\noindent \textbf{Human-Agent Interaction Quality.}
Interaction is the core process of LLM-HAS, and its quality directly influences the final outcome. This evaluation should encompass key aspects such as naturalness, coherence, seamlessness, and readability. Such criteria have been widely adopted in the evaluation of LLM-based agents in collaborative tasks, including MEGANno+ ~\citep{kim2024meganno+}, tAIfa ~\citep{almutairi2025taifa}, and studies on AI awareness ~\citep{cheng2025exploratory}.

\noindent \textbf{Trust, Transparency, and Explainability.} LLM-HAS should be evaluated across the domains of trust, transparency, and explainability, as these factors directly influence human decision when system outputs are applied to real-world tasks. Key evaluation criteria within this domain include: accuracy and correctness, explainability and transparency, perceived competence, benevolence, integrity, system reliability, and user control. These dimensions have been extensively studied even prior to the rise of LLM-HAS~\citep{khavas2021review}. More recently, research has focused on trust in LLMs in contexts such as LLM-based planning systems ~\citep{chen2025evaluating}, question-answering tasks ~\citep{ding2025citations}, and human-LLM collaboration scenarios ~\citep{10.1145/3491102.3517582}.

\noindent \textbf{Ethical Alignment and Safety.} 
Given the powerful capabilities of LLM-HAS, it is essential to ensure they are directed toward beneficial and responsible use. Key evaluation aspects of this criteria include robustness, misuse prevention, operational safety, and protection of privacy and security. This domain has gained significant attention since the rise in popularity of LLMs ~\citep{askell2021general}. 

\noindent \textbf{User Experience and Cognitive Load.} User Experience (UX) is a particularly important and distinctive evaluation criterion for LLM-HAS. In these systems, UX involves the full spectrum of user perceptions, emotional responses, satisfaction, and overall impressions. \cite{dierk2025evaluating} identified three primary strategies for evaluating UX in this context: Assessing the direct outputs generated by LLMs, evaluating co-created artifacts produced through human-agent collaboration, and analyzing user subjective experiences during their interaction with the system.

\section{Alternative Views}\label{Section-7}

\subsection{A Framework for Evaluating Human-Agent Systems}
Before addressing specific critiques of LLM-HAS, we introduce an abstract framework to better understand the underlying trade-offs.

\begin{definition}[Utility of Agent Systems]\label{def:utility}
The utility ($U$) of an agent system can be represented as:
\begin{equation}\label{eq:utility}
U = V \cdot S - C_h - C_e
\end{equation}
where $V$ represents the value created per successful task, $S$ is the success rate (0-1), $C_h$ denotes human costs, and $C_e$ encompasses error costs (including financial, reputational, and societal costs).
\end{definition}

\begin{definition}[Optimal Human Involvement]\label{def:optimal}
The optimal degree of human involvement $h^*$ is the level that maximizes the system's utility:
\begin{equation}\label{eq:optimal}
h^* = \arg\max_h U(h)
\end{equation}
This optimal point balances the increased success rate and reduced error costs against the additional human operational costs.
\end{definition}

\subsection{The Enduring Appeal of Fully Autonomous Agents}
\textbf{View:} The vision of Fully Autonomous Agents (FAA) remains compelling for several reasons. Proponents often highlight the potential for significant cost reductions through the automation of human labor, substantial increases in speed and operational efficiency for various tasks, the ability to scale operations rapidly without commensurate increases in human resources, and the capacity for continuous 24/7 operation without issues like fatigue or breaks.

\textbf{Response:} Within our framework from Definition~\ref{def:utility}, FAA systems typically minimize $C_h$ but may significantly increase $C_e$ due to higher error rates or more severe consequences when errors occur. We acknowledge that as AI capabilities advance, the optimal human involvement $h^*$ defined in Equation~\ref{eq:optimal} will gradually decrease. However, current technological limitations mean that the optimal balance still requires significant human participation.

% A2: 人类参与成为瓶颈
\subsection{Human Involvement as a Bottleneck - Low Quality and Unreliability of Human Feedback}
% A2-1: 人类参与成为瓶颈: 低质量的人类反馈
% \subsubsection{Low Quality and Unreliability of Human Feedback.}
\textbf{View:} Humans feedback can often be noisy, biased, inconsistent, or even incorrect. 
Such imperfect feedback could degrade LLM agent performance or introduce new problems.
Therefore, it might be preferable for agents to learn from more structured data sources or through self-play mechanisms.

\textbf{Response:} It is true that human feedback is not infallible. However, a core focus of LLM-HAS research is developing mechanisms to efficiently elicit and integrate high-quality feedback. Even imperfect human input often contains contextual knowledge and domain-specific insights that are currently absent in LLMs. In our conceptual framework from Equation~(\ref{eq:utility}), while suboptimal human feedback may temporarily decrease system success rate ($S$), the alternative—an absence of human feedback—can lead to agents that perpetuate their own biases without correction mechanisms, potentially leading to much higher error costs ($C_e$). The goal of LLM-HAS is to harness the strengths of human input while mitigating its weaknesses through careful system design, thereby approaching the optimal involvement level $h^*$ defined in Equation~(\ref{eq:optimal}).

\vspace{-1mm}

% A2-2: 人类参与成为瓶颈: 响应性与延迟
\subsection{Human Involvement as a Bottleneck - Responsiveness and Delay}

\textbf{View:} Requiring human intervention or feedback will inevitably slow down the system's response time, making it unsuitable for time-critical applications. The ``human in the loop'' can become the ``human as a bottleneck,'' negating the speed advantages of AI.

\textbf{Response:} This concern is valid for certain time-critical applications. However, within our utility model in Definition~\ref{def:utility}, response time is one component of the overall value equation. For many complex cognitive tasks, a slight delay to incorporate human input is acceptable if it significantly improves the success rate ($S$). LLM-HAS designs can optimize when to solicit human help, performing many subtasks autonomously and involving humans only at critical decision points. This strategic integration of human expertise often yields higher total utility ($U$) despite some speed trade-offs. As AI capabilities advance, the optimal human involvement $h^*$ will likely decrease, further improving response times while maintaining high success rates.

\subsection{The Human Cost Factor in HAS}
\textbf{View:} Integrating humans into the operational loop (e.g., for providing feedback, oversight, or direct collaboration) is expensive in terms of human time, effort, and the need for specialized training. This can negate the economic benefits that AI automation is expected to deliver.

\textbf{Response:} The human operational costs ($C_h$) must be weighed against the potentially far greater error costs ($C_e$) associated with failures in fully autonomous systems. In high-stakes domains, error costs can include accidents, reputational damage, legal liabilities, and loss of user trust. Effective LLM-HAS design aims to optimize human effort, focusing it where it adds the most value. Furthermore, the value ($V$) delivered by an LLM-HAS that successfully tackles a complex problem can far outweigh the operational cost of human collaboration. As AI capabilities improve, we expect the optimal degree of human involvement $h^*$ from Definition~\ref{def:optimal} to decrease, reducing $C_h$ while maintaining high success rates. This evolution represents a responsible path toward increasingly autonomous systems that optimize the total utility function rather than simply minimizing human costs at the expense of other factors.

\section{Conclusion}\label{Section-8}
This position paper calls for a strategic shift from aggressively pursuing full autonomy to prioritizing LLM-based human-agent systems at this developmental stage. Despite significant advancements inspired by recent breakthroughs in LLM technology, the premature and widespread deployment of fully autonomous agents presents critical risks related to reliability, complexity, and legal issues across diverse application domains. We support this position by building on the concept of the human-agent system and detailing its practical development roadmap, design principles, and key challenges. In addition, we provide detailed implementation guidelines to help the AI research community effectively embrace, evaluate, and collaboratively advance this emerging paradigm.

\bibliographystyle{plain}
\bibliography{custom.bib}

%%%%%%%%%%%%%%%%%%%%%%%%%%%%%%%%%%%%%%%%%%%%%%%%%%%%%%%%%%%%

% \appendix

% \section{Technical Appendices and Supplementary Material}
% Technical appendices with additional results, figures, graphs and proofs may be submitted with the paper submission before the full submission deadline (see above), or as a separate PDF in the ZIP file below before the supplementary material deadline. There is no page limit for the technical appendices.

\end{document}